\documentclass{midl} 

\usepackage{multirow}

\usepackage{mwe} 
\jmlrvolume{-- Under Review}
\jmlryear{2022}
\jmlrworkshop{Full Paper -- MIDL 2022 submission}
\editors{Under Review for MIDL 2022}

\title[Short Title]{Representative Image Feature Extraction via Contrastive Learning Pretraining for Chest X-ray Report Generation}

\midlauthor{\Name{Yu-Jen Chen\midljointauthortext{Contributed equally}\nametag{$^{1}$}} \Email{yujenchen@gapp.nthu.edu.tw}\\
\Name{Wei-Hsiang Shen\midlotherjointauthor\nametag{$^{2}$}} \Email{whshen@gapp.nthu.edu.tw}\\
\Name{Hao-Wei Chung\nametag{$^{1}$}} \Email{xdmanwww@gapp.nthu.edu.tw}\\
\Name{Ching-Hao Chiu\nametag{$^{1}$}} \Email{gwjh101708@gapp.nthu.edu.tw}\\
\Name{Da-Cheng Juan\nametag{$^{1}$}} \Email{dacheng@gapp.nthu.edu.tw}\\
\Name{Tsung-Ying Ho\nametag{$^{3}$}} \Email{albertyho@gmail.com}\\
\Name{Chi-Tung Cheng\nametag{$^{4}$}} \Email{atong89130@gmail.com}\\
\Name{Meng-Lin Li\nametag{$^{2}$}} \Email{mlli@ee.nthu.edu.tw}\\
\Name{Tsung-Yi Ho\nametag{$^{1}$}} \Email{tyho@cs.nthu.edu.tw}\\
\addr $^{1}$ Department of Computer Science, National Tsing Hua Unviserity, Taiwan \\
\addr $^{2}$ Department of Electrical Engineering, National Tsing Hua Unviserity, Taiwan \\
\addr $^{3}$ Department of Nuclear Medicine, Chang Gung Memorial Hospital, Linkou, Taiwan \\
\addr $^{4}$ Department of Trauma and Emergency Surgery, Chang Gung Memorial Hospital, Linkou, Taiwan \\
}

\begin{document}

\maketitle

\begin{abstract}
Medical report generation is a challenging task since it is time-consuming and requires expertise from experienced radiologists. The goal of medical report generation is to accurately capture and describe the image findings. Previous works pretrain their visual encoding neural networks with large datasets in different domains, which cannot learn general visual representation in the specific medical domain. In this work, we propose a medical report generation framework that uses a contrastive learning approach to pretrain the visual encoder and requires no additional meta information. In addition, we adopt lung segmentation as an augmentation method in the contrastive learning framework. This segmentation guides the network to focus on encoding the visual feature within the lung region. Experimental results show that the proposed framework improves the performance and the quality of the generated medical reports both quantitatively and qualitatively.

\end{abstract}


\section{Introduction}
Medical Images, such as X-rays, have been widely used for supporting clinical diagnosis and diseases treatment.
Radiologists are responsible for delivering the interpretation and diagnosis based on their medical expertise and providing the medical reports as reference. 
However, medical report generation is time-consuming and requires medical expertise from experienced radiologists, which may not be abundant in rural area. Thus, automatic medical report generation is essential in reducing workload, facilitating clinical workflow, and providing additional diagnostic insights.



Most of the current state-of-the-art methods are based on encoder-decoder frameworks \cite{shin2016learning}. 
Many of them \cite{jing2017automatic,li2018hybrid,chen2020generating} utilize pretrained encoders to extract image features and design decoders to generate the corresponding reports.
Although these works design new architectures which result in significant results, their encoders remain to be pretrained with datasets of different domains, where some medical characteristics cannot be fully learned and captured, and lead to limited performance improvement.

In this work, we proposed a medical report generating framework by leveraging the power of Contrastive Learning (CL) and lung segmentation without using extra datasets and additional metadata such as keywords in our model.
The encoder is first pretrained with CL using images only and then the entire network is trained end-to-end under a supervised setting (images and reports).
Experiments showed that the representation trained by the CL framework results in better performance than other commonly used pretraining settings. In addition, in the CL model, we propose a new augmentation method, lung segmentation, which helps the model attend to the lung region of the given images. The lung segmentation augmentation guides the model to focus on lung regions and generate more detailed and accurate descriptions around the lung regions.
In summary, we demonstrate that CL with lung segmentation improves the performance and the quality of the generated medical reports. Moreover, applying the CL pretrained encoder to state-of-the-art methods also boosts their performance.

\section{Related Work}
\subsection{Chest X-ray Report Generation}
Most existing report generating approaches \cite{yuan2019automatic,syeda2020chest,xue2018multimodal,li2020vispi} follow the traditional encoder-decoder architecture to design their model.
\cite{jing2017automatic} proposed a powerful model combining the co-attention mechanism and the multi-level LSTM to generate a longer paragraph of reports. 
The co-attention mechanism will combine the visual feature and the paragraph information to achieve better performance. 
\cite{li2018hybrid} harnessed the reinforcement-learning strategy to update the HRGR-Agent via sentence-level and word-level rewards and maintained a high accuracy in report diagnosis. Finally, \cite{chen2020generating} proposed a memory-based Transformer model to explore the correlation between  similar images in the dataset, and such model can report necessary medical terms as well as a meaningful attention map.
However, these works focus on designing novel decoders and pretrain their encoders with datasets of domains not related to the application under concern. The medical characteristic cannot be fully learned and captured with such encoder settings. 
In comparison, our results show that our training framework provides better-encoded features that are applicable to state-of-the-art methods.



\subsection{Contrastive Learning (CL)}
Contrastive learning has shown great promise to learn image representation via reflecting the difference between image features and boosting the performance of various image-related tasks.
In fundamental CL frameworks, such as SimCLR \cite{chen2020simple}, MoCo \cite{he2020momentum}, and PIRL \cite{Misra_2020_CVPR}, the feature extractor is trained by minimizing the contrastive loss of the two positive related images and repelling the two negative related images. 
In the medical domain, several previous works \cite{chen2020two,liu2021contrastive,liu2019align} also explore the capability that the CL framework achieve performance improvement for medical images.

\section{Methodology}


Most existing automatic chest X-ray report generation approaches follow an encoder-decoder framework, which consists of an image encoder and a report decoder. 
The image encoder $E(\cdot) = I \rightarrow V$  encodes the input image $I$ into the visual representation $V$ in latent space. The report decoder $D(\cdot) = V \rightarrow R$ is used to generate the report $R$ from the given image representation $V$. ResNet is often used as the encoder $E$ and Transformer is often used as the decoder $D$. The encoder-decoder framework is trained in a supervised manner in which paired chest X-ray images and their corresponding medical reports are required.

In this work, we explore the importance of effective visual representation for chest X-ray report generation. Our image encoder network $E$ is pretrained with contrastive learning techniques that incorporate lung segmentation as one of the augmentation methods.


\subsection{Contrastive Learning (CL) Pretraining}
\label{sec:cl}
In this work, different from most existing approaches whose image encoders are pretrained with large datasets with noisy labels, we adopt the Contrastive Learning (CL) approaches SimCLR \cite{chen2020simple} and MoCo \cite{he2020momentum} to pretrain the image encoder. 

\begin{figure}[h]
\centering
\begin{minipage}{0.55\textwidth}
  \centering
\includegraphics[width=1.0\textwidth]{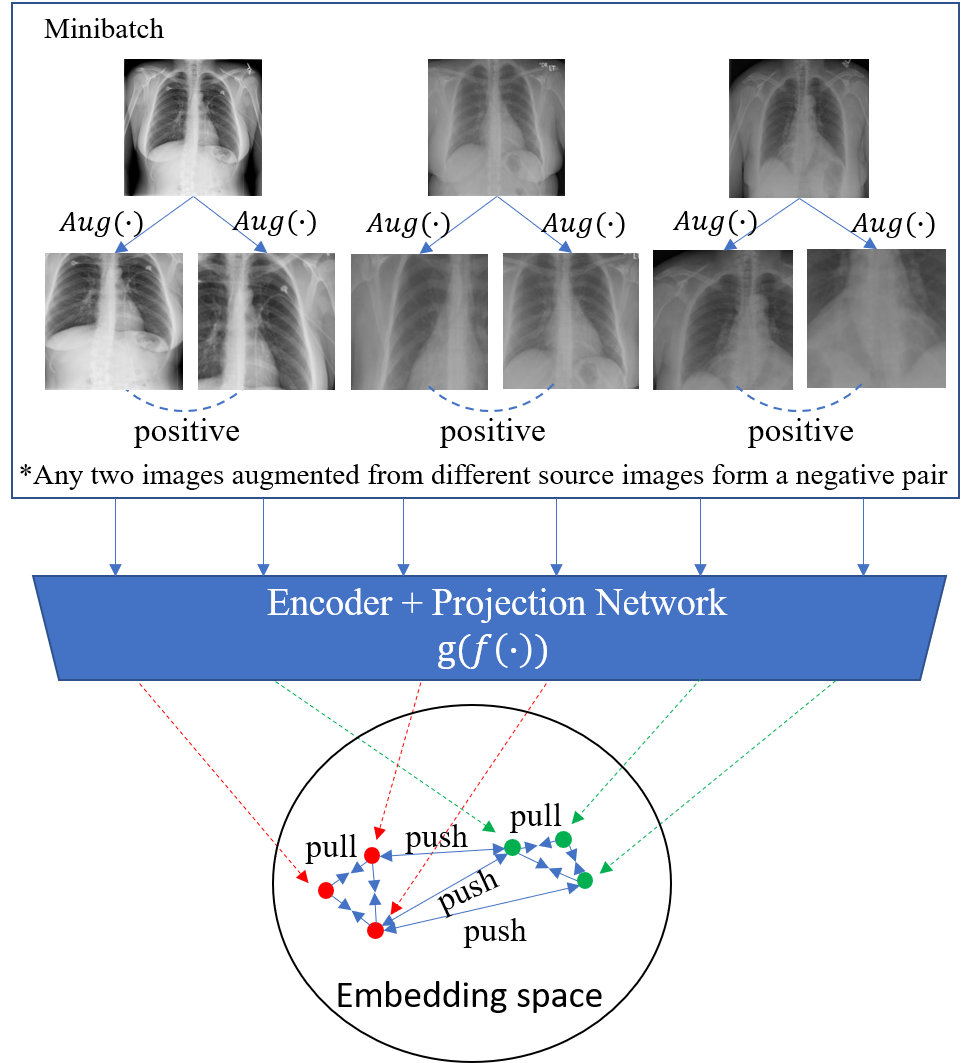}
\end{minipage}%
\begin{minipage}{0.45\textwidth}
  \centering
\includegraphics[width=1.0\textwidth]{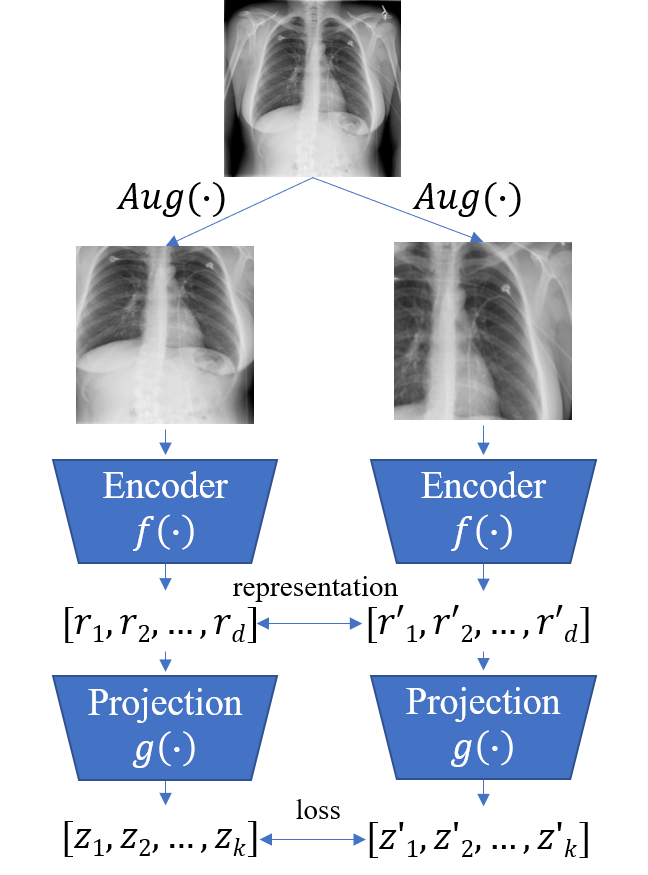}
\end{minipage}%
\caption{An illustration of the fundamental concept for CL. The left figure shows how positive and negative image pairs are created. The figure on the right shows the details of the network, which is a combination of an encoder and a projection network.} \label{fig:cl}
\end{figure}


CL is to minimize the image embedding distance between ``positive'' image pairs, and maximize the distance between ``negative'' image pairs. A positive image pair is created by applying two different augmentations to a single image, and a negative image pair is two augmented images that come from two different source images. In this work for X-ray images, the augmentation techniques used are random resized crop and random horizontal flip.

The augmented images are then passed to an encoder neural network $f(\cdot)$, and encoded to their representation vectors $r=f(x)\in R^d$, where $d$ is the dimension of the representation vector, as shown in Fig. \ref{fig:cl}. Then the representation vectors are projected to a fewer dimension space $z=g(r)\in R^k$ by a project network $g(\cdot)$, which is a neural network of two fully-connected layers.





Finally, for a minibatch of $N$ images, let $i\in I\equiv\{1, 2, ..., 2N\}$ be the index of the augmented sample in the minibatch, the contrastive loss function is formulated as follows:
\begin{equation}
\large
    L = -\sum_{i\in I} \frac{\exp(z_i \cdot z_{j(i)}/\tau)}{\sum_{k\in B(i)} \exp(z_i \cdot z_k/\tau)}
\end{equation}
where $z_i=g(f(x))\in R^k$ is the projected vector of the image $x$, the $\cdot$ operator is the inner product, index $j(i)$ is the positive image of anchor $i$, $\tau = 0.5$ is the temperature parameter, and $B(i)\equiv I \setminus \{i\}$. The numerator of the loss calculates all the vector similarity of positive pairs, and the denominator is the summation of similarity of all the image pairs (except the positive pairs) inside the minibatch.

\subsection{Lung Segmentation Contrastive Learning}
\label{sec:segssl}

\begin{figure}
    \centering
    \includegraphics[width=0.9\linewidth]{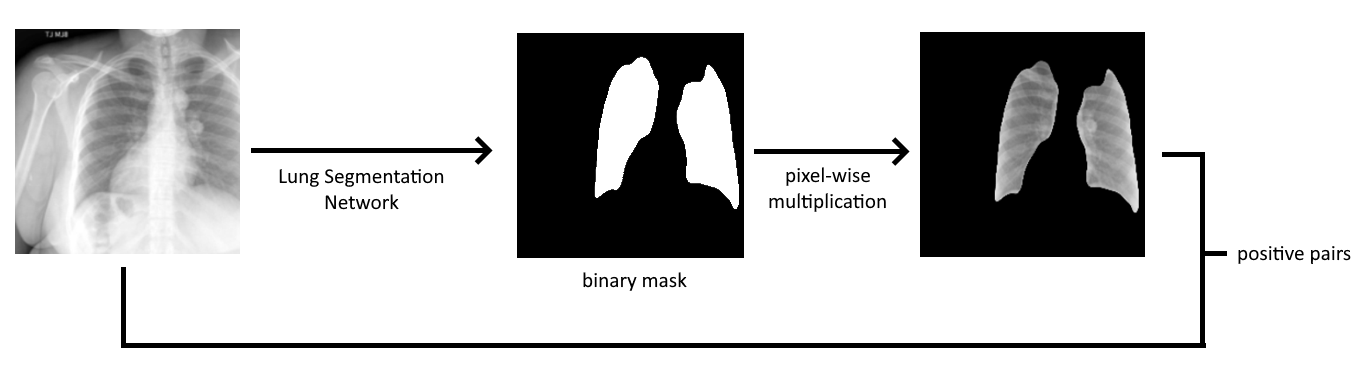}
    \caption{An example of positive pair with the lung masking augmentation added. We can see that the images in a positive pair contain lung image and whole image.}
    \label{fig:lung}
\end{figure}

In addition to conventional image augmentation techniques, we explore segmentation methods as augmentation methods to support the model to generate medical findings related to the segmented region.
We use a trained lung segmentation model\footnote{\url{https://github.com/rani700/xray}} as an augmentation function, which guides the network by focusing on the lung region. The segmentation model is randomly applied with the probability of 0.5 and the binary segmentation mask crops out the lung region in the image, as shown in Fig. \ref{fig:lung}. Since the contrastive learning setting guides the network to minimize the distance between images with and without lung segmentation crop-off, the network learns to focus less on the non-lung region and generates more detailed features of the lung region. Thus, the report decoder receives lung-specific features and generates reports associated with tissue or abnormality of the lung or rib.


\section{Experimental Settings}
\subsection{Dataset}
\label{sec:dataset}


In this work, we implement report generation on two datasets, Indiana University Chest X-ray Collection (IU X-ray) dataset \cite{demner2016preparing} which contains 7,470 pairs of frontal and lateral chest X-rays and reports, and the Medical Information Mart for Intensive Care - Chest X-ray (MIMIC-CXR) dataset \cite{johnson2019mimic}, which contains 377,110 images and 227,835 radiographic reports. We duplicated the data-preprocess of \cite{chen2020generating} and followed their dataset split on both dataset, which is widely adopted by many related works \cite{jing2017automatic,chen2020generating,rennie2017self,lu2017knowing,li2018hybrid}. In detail, IU X-ray dataset is divided by
7:1:2 division of the entire dataset into training/validation/testing sets, as for MIMIC-CXR, we use the official split.

The experiments are evaluated with quantitative NLP metrics that measure the similarity between the predicted and target sentence sequence. The metrics used include BLEU \cite{10.3115/1073083.1073135}, METEOR \cite{denkowski:lavie:meteor-wmt:2014}, and ROGUE-L \cite{lin-2004-rouge}. BLEU measures the similarity between the two sentences by calculating the number of word matches in a weighted manner. These matches are independent of position. METEOR is similar to BLEU but uses the harmonic mean of unigram precision and recall, with recall weighted higher than precision. 
ROGUE-L takes into account the structural similarity at the sentence level and identifies the longest co-occurring n-grams in the sequence.

\subsection{Implementation Details}

We adopt the ResNet-50 architecture as the backbone of the encoder in all experiments. At the pretraining phase for the CL approaches (SimCLR and MoCo), all the images in the training set are used, and the model is trained for 500 epochs with the batch size as 64. As for the decoder training (fine-tuning step), the images and the corresponded report in the training set are used, and the entire framework is trained for 200 epochs with the batch size set as 64. Adam optimizer with the learning rate set as 0.001 and decay to 0.00001 followed by the cosine scheduler. We also use the weight decay of 0.000001 to avoid the overfitting problem. The validation set and the test set are used only for evaluation.

\section{Results and Discussion}


\subsection{Comparison across Different Encoder Settings}

In this section, we compared pretraining methods for the encoder to show the feasibility of the proposed framework. 
The compared pretraining methods include using AutoEncoder (AE) pretraining \cite{baldi2012autoencoders}, Imagenet dataset (Imagenet) pretraining, Multi-Label Classification (MLC) pretraining \cite{bi2013efficient}, without pretraining (scratch), MoCo pretraining, SimCLR pretraining, and SimCLR pretraining with lung segmentation included for the encoder. The labels used in the MLC pretraining are the MeSH (Medical Subject Headings) tag extracted from the reports (the MIMIC-CXR dataset did not provide MeSH tags for the reports). In this section, we discuss using Transformer, LSTM, and GRU as the decoder architecture. 
The pretrained weight of the encoder is assigned and trained end-to-end with the decoder.

\begin{table}[h]
\centering
\small
\begin{tabular}{cc|rrrrrr}
\hline
\multicolumn{1}{c|}{Dataset}                      & \multicolumn{1}{c|}{Pretraining Method}        & \multicolumn{1}{c}{B-1} & \multicolumn{1}{c}{B-2} & \multicolumn{1}{c}{B-3} & \multicolumn{1}{c}{B-4} & \multicolumn{1}{c}{M} & \multicolumn{1}{c}{R-L} \\ \hline
\multicolumn{1}{c|}{\multirow{7}{*}{IU X-ray}}    & Scratch             & 0.404                   & 0.249                   & 0.172                   & 0.126                   & 0.161                 & 0.320         \\
\multicolumn{1}{c|}{}                             & AE                  & 0.417                   & 0.273                   & 0.195                   & 0.148                   & \textbf{0.176}        & 0.361         \\
\multicolumn{1}{c|}{}                             & Imagenet            & 0.420                   & 0.256                   & 0.175                   & 0.127                   & 0.167                 & 0.326         \\
\multicolumn{1}{c|}{}                             & MLC                 & 0.423                   & 0.270                   & 0.190                   & 0.141                   & \textbf{0.176}        & 0.359         \\
\cline{2-8}
\multicolumn{1}{c|}{}                             & MoCo                & 0.426                   & 0.259                   & 0.175                   & 0.126                   & 0.168                 & 0.325         \\
\multicolumn{1}{c|}{}                             & SimCLR              & \textbf{0.456}          & \textbf{0.289}          & \textbf{0.202}          & \textbf{0.150}          & \textbf{0.176}        & \textbf{0.362}\\
\multicolumn{1}{c|}{}                             & SimCLR (w/ Lung Seg)& 0.448                   & 0.282                   & 0.195                   & 0.141                   & \textbf{0.176}        & 0.340         \\
\hline
\multicolumn{1}{c|}{\multirow{6}{*}{MIMIC-CXR}}   & Scratch             & 0.211                   & 0.132                   & 0.090                   & 0.064                   & 0.105                 & 0.250         \\
\multicolumn{1}{c|}{}                             & AE                  & 0.316                   & 0.186                   & 0.119                   & 0.081                   & 0.119                 & 0.243         \\
\multicolumn{1}{c|}{}                             & Imagenet            & 0.270                   & 0.166                   & 0.109                   & 0.076                   & 0.114                 & 0.255         \\
\cline{2-8}
\multicolumn{1}{c|}{}                             & MoCo                & 0.334                   & 0.203                   & 0.133                   & 0.091                   & 0.129                 & 0.261         \\
\multicolumn{1}{c|}{}                             & SimCLR              & \textbf{0.343}          & \textbf{0.208}          & \textbf{0.137}          & \textbf{0.095}          & \textbf{0.132}        & 0.260         \\
\multicolumn{1}{c|}{}                             & SimCLR (w/ Lung Seg)& 0.325                   & 0.198                   & 0.130                   & 0.090                   & 0.129                 & \textbf{0.265}\\
\hline
\end{tabular}
\caption{Quantitative comparison among different encoders when using Transformer as decoder. B-n, M, and R-L are short for BLEU-n, METEOR, and ROUGE-L, respectively.}
\label{Table:Transformer_dif_encoder}
\end{table}

\begin{figure}[h]
    \centering
    \includegraphics[width=1\linewidth]{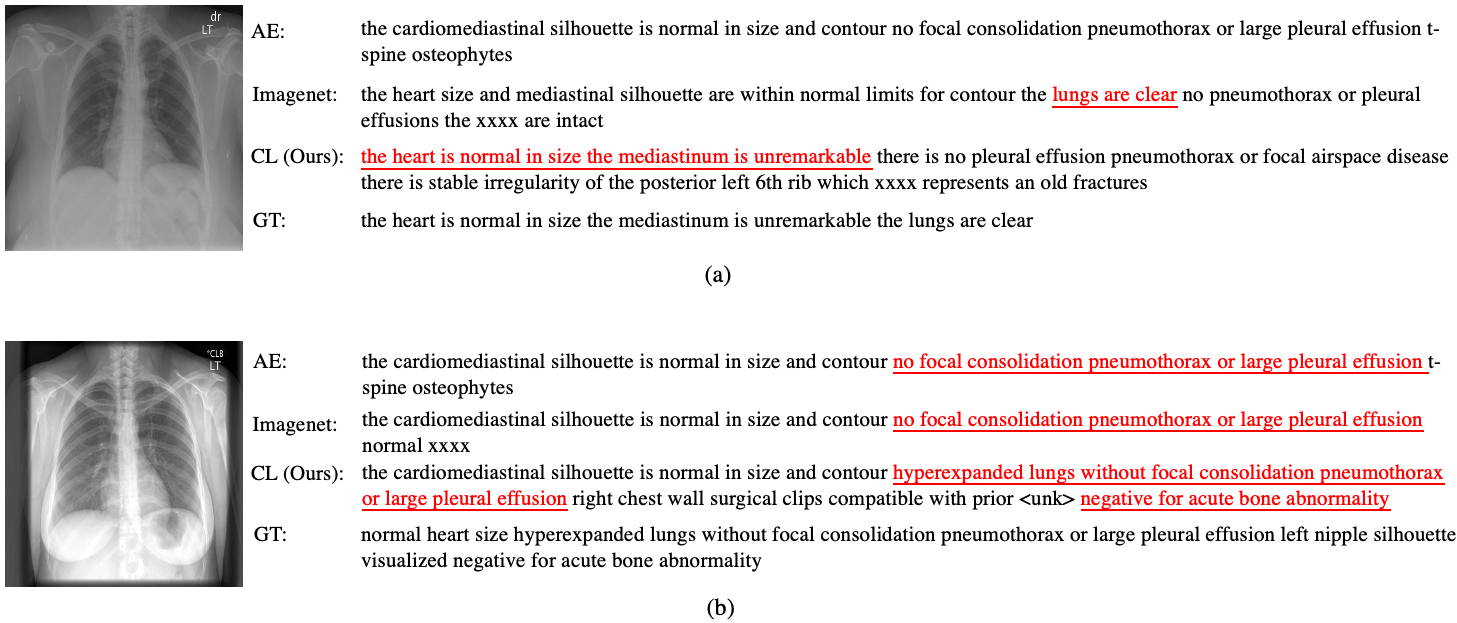}
    \caption{Qualitative comparison among different encoders when using Transformer as the decoder. The descriptions marked with red underline represent to match the ground truth. AE and GT stand for the Auto-encoder pretraining and the ground truth, respectively.}
    \label{fig:Transformer-cl-encoder}
\end{figure}

\subsubsection{Transformer Quantitative Comparison} As shown in Table \ref{Table:Transformer_dif_encoder}, in the IU X-ray dataset, the one without pretraining leads to the worst results, which is expected since the encoder has no prior knowledge of the image. Afterward, due to the expensive pixel-wise image reconstruction, it is difficult for AE to learn the details of images, especially medical images. 
As for the weight pretrained by Imagenet, though it contains significant information of the nature images, there is a domain gap from chest X-ray images. In MLC pretraining, keywords are extracted from the report, and the model is pretrained with the supervised multi-label classification task. As the encoder has learned little prior knowledge of the reports, we observe that the results are comparable with CL pretraining. Ultimately, CL (SimCLR and MoCo) pretraining outperforms all other pretraining strategies in almost all the metrics. For example, the performance boost at BLEU-1 score compared with other pretraining strategies are up to 9.0\%, and 2.25\% for SimCLR and MoCo, respectively.

As for the MIMIC-CXR dataset, as shown in Table \ref{Table:Transformer_dif_encoder}, the improvement trend is similar to that of IU X-ray dataset. CL (SimCLR and MoCo) pretraining outperforms all other pretraining strategies in all the metrics. The performance boost at BLEU-1 score compared with other pretraining strategies is up to 32.0\%, and 28.0\% for SimCLR and MoCo, respectively.



\subsubsection{Transformer Qualitative Comparison} Fig. \ref{fig:Transformer-cl-encoder} shows the ground truth and the predicted description by different encoders. In case (a), the AE pretrained model generates the description which did not match the ground truth, and the Imagenet pretrained model only describes that the lung is clear, whereas, with CL model, it points out the heart condition and mediastinum condition. In case (b), only CL pretrained model points out ``hyperexpanded lungs'' and ``negative for acute bone abnormality'', whereas the Imagenet pretrained model and the AutoEncoder predict some common information. This shows that the CL pretrained encoder generates precise descriptions when using the Transformer decoder.

\begin{table}[!h]
\centering
\small
\begin{tabular}{cc|rrrrrr}
\hline
\multicolumn{1}{c|}{Dataset}                      & \multicolumn{1}{c|}{Pretraining Method}        & \multicolumn{1}{c}{B-1} & \multicolumn{1}{c}{B-2} & \multicolumn{1}{c}{B-3} & \multicolumn{1}{c}{B-4} & \multicolumn{1}{c}{M} & \multicolumn{1}{c}{R-L} \\ \hline
\multicolumn{1}{c|}{\multirow{6}{*}{IU X-ray}}    & Scratch  & 0.299                   & 0.185                   & 0.132                   & 0.100                   & 0.134                 & 0.324                   \\
\multicolumn{1}{c|}{}                             & AE       & 0.389                   & 0.233                   & 0.157                   & 0.111                   & 0.154                 & 0.317                   \\
\multicolumn{1}{c|}{}                             & Imagenet & 0.397                   & 0.246                   & 0.170                   & 0.126                   & 0.160                 & \textbf{0.329}          \\
\multicolumn{1}{c|}{}                             & MLC      & 0.410                   & 0.250                   & 0.170                   & 0.123                   & 0.164                 & 0.325                   \\
\cline{2-8}
\multicolumn{1}{c|}{}                             & MoCo                & 0.425                   & 0.260                   & 0.177                   & 0.128                   & 0.169        & 0.326         \\
\multicolumn{1}{c|}{}                             & SimCLR              & 0.426          & 0.265          & 0.181          & 0.131   & 0.168                 & 0.326    \\
\multicolumn{1}{c|}{}                             & SimCLR (w/ Lung Seg)&  \textbf{0.431}   & \textbf{0.268}  & \textbf{0.185}  & \textbf{0.137}                    & \textbf{0.172}       & \textbf{0.332}         \\
\hline

\multicolumn{1}{c|}{\multirow{5}{*}{MIMC-CXR}}    & Scratch  & 0.312                   & 0.193                   & 0.130                   & 0.092                   & 0.128                 & 0.272                   \\
\multicolumn{1}{c|}{}                             & AE       & 0.282                   & 0.174                   & 0.117                   & 0.083                   & 0.119                 & 0.264                   \\
\multicolumn{1}{c|}{}                             & Imagenet & 0.309                   & 0.193                   & 0.131                   & 0.094                   & 0.130                 & 0.276                   \\
\cline{2-8}
\multicolumn{1}{c|}{}                             & MoCo                & 0.309                  & 0.191                    & 0.128                   & 0.091                   & 0.127        & 0.270         \\
\multicolumn{1}{c|}{}                             & SimCLR              & \textbf{0.330}          & \textbf{0.206}          & \textbf{0.139}          & \textbf{0.099}   & \textbf{0.134}                 & \textbf{0.278}    \\
\multicolumn{1}{c|}{}                             & SimCLR (w/ Lung Seg)&  0.325   & 0.203  & 0.138  & 0.095                    & \textbf{0.134}                 & \textbf{0.278}         \\
\hline
\end{tabular}
\caption{Quantitative comparison among different encoders when using LSTM as decoder. B-n, M, and R-L are short for BLEU-n, METEOR, and ROUGE-L, respectively.}
\label{Table:LSTM_dif_encoder}
\end{table}

\begin{figure}[!h]
    \centering
    \includegraphics[width=1\linewidth]{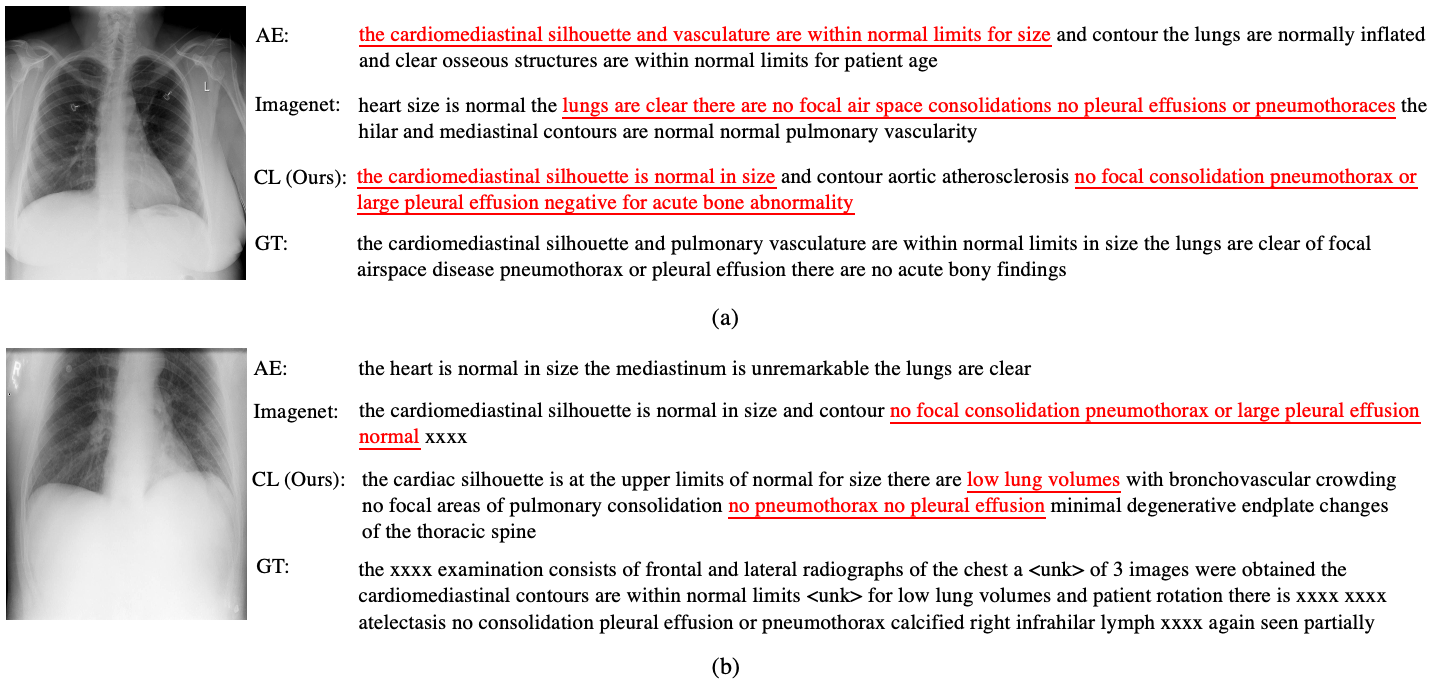}
    \caption{Qualitative Comparison among Different Encoders when using LSTM as the decoder. The descriptions marked with red underline represent ground truth matching. AE and GT stand for the Auto-encoder preatraining and the ground truth, respectively.}
    \label{fig:lstm-cl-encoder}
\end{figure}

\subsubsection{LSTM Quantitative Comparison}
Table \ref{Table:LSTM_dif_encoder} shows that SimCLR pretraining improves the BLEU-1 and BLEU-4 score up to 42\% and 31\%, respectively, compared to other pretraining strategies on the IU X-ray dataset. Moreover, in the MIMIC-CXR dataset, CL pretraining outperforms other pretraining strategies in all metrics.

\subsubsection{LSTM Qualitative Comparison}
Fig. \ref{fig:lstm-cl-encoder} shows the ground truth and the predicted description by different encoders when using LSTM as a decoder. In case (a), the AutoEncoder pretrained model only matches the mediastinum description with GT, and the Imagenet pretrained model describes the lung parenchyma and pleura condition, whereas, with CL model, it points out all of them with the bone condition additionally.  In case (b), only the CL pretrained model points out ``low lung volumes'', whereas the Imagenet pretrained model only predicts that the lung stays normal. This shows that the CL encoder aids the model in generating precise descriptions.

\subsubsection{GRU Quantitative Comparison}
Table \ref{Table:GRU_dif_encoder} shows that in the IU X-ray dataset, SimCLR and MoCo pretraining boost the BLEU-1 score up to 7.75\% and 7.74\%, respectively, compared with other pretraining strategies. As for the MIMIC-CXR dataset, SimCLR pretraining achieves the highest score in all evaluate metrics, about 5.5\% higher than without pretraining the encoder (Scratch).

\subsubsection{GRU Qualitative Comparison}
\begin{figure}[!h]
    \centering
    \includegraphics[width=1\linewidth]{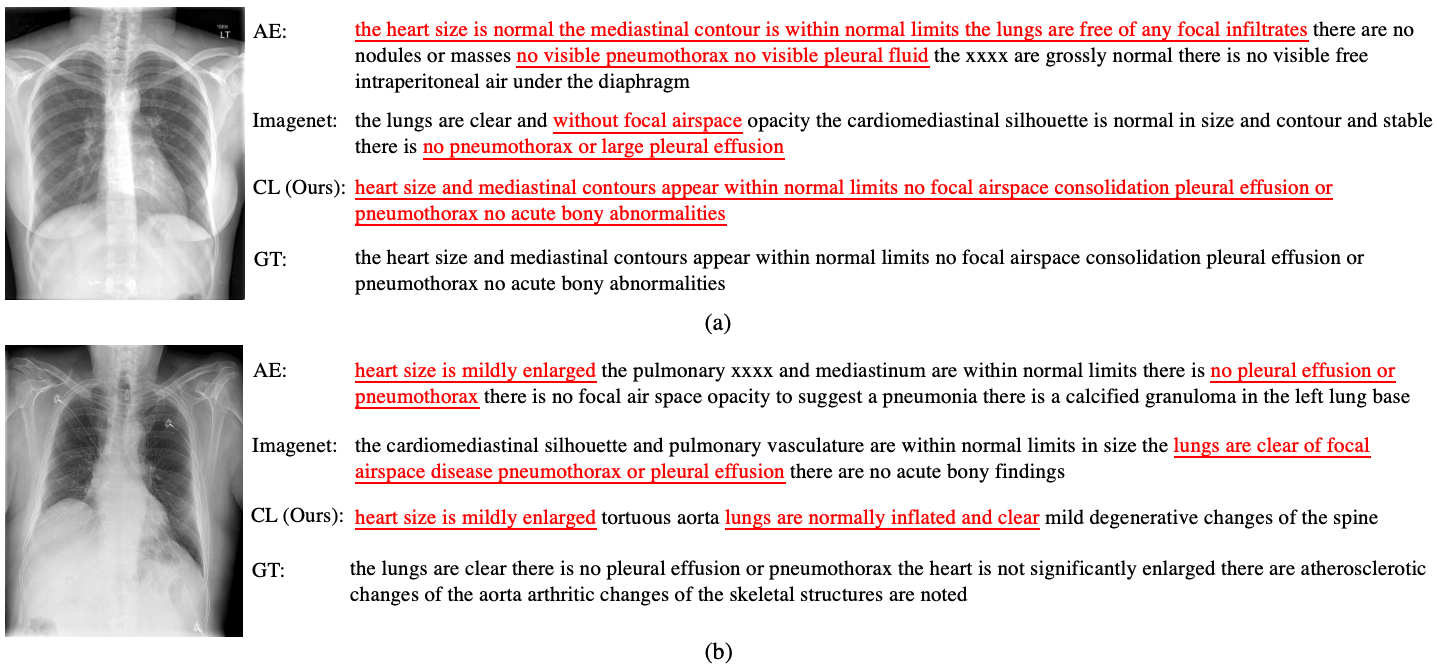}
    \caption{Qualitative comparison among different encoders when using GRU as the decoder. The descriptions marked with red underline represent to match the ground truth. AE and GT stand for the Auto-encoder preatraining and the ground truth, respectively.}
    \label{fig:GRU-cl-encoder}
\end{figure}

Fig. \ref{fig:GRU-cl-encoder} shows the ground truth and the predicted description by different encoders when using GRU as the decoder. In case (a), the Auto-Encoder pretrained model did not describe the bone condition, and the Imagenet pretrained model only describes the lung condition, whereas the CL pretrained model points out both of them.  In case (b), only CL pretrained model point out ``heart is mildly enlarged'' and the lung had no problem, whereas the Imagenet pretrained model only predicts that the lung stays normal, and the AutoEncoder predicts others false information. Qualitatively, the CL encoder aids the model in generating precise descriptions while using the GRU decoder than other pretraining approaches.

\begin{table}[!h]
\centering
\small
\begin{tabular}{cc|rrrrrr}
\hline
\multicolumn{1}{c|}{Dataset}                      & \multicolumn{1}{c|}{Pretraining Method}        & \multicolumn{1}{c}{B-1} & \multicolumn{1}{c}{B-2} & \multicolumn{1}{c}{B-3} & \multicolumn{1}{c}{B-4} & \multicolumn{1}{c}{M} & \multicolumn{1}{c}{R-L} \\ \hline
\multicolumn{1}{c|}{\multirow{7}{*}{IU X-ray}}    & Scratch  & 0.351                   & 0.221                   & 0.156                   & 0.117                   & 0.153                 & 0.333                   \\
\multicolumn{1}{c|}{}                             & AE       & 0.389                   & 0.241                   & 0.166                   & 0.121                   & 0.161                 & 0.332                   \\
\multicolumn{1}{c|}{}                             & Imagenet & 0.417                   & 0.254                   & 0.173                   & 0.124                   & 0.165                 & 0.328                   \\
\multicolumn{1}{c|}{}                             & MLC      & 0.412                   & \textbf{0.263}          & \textbf{0.189}          & \textbf{0.145}          & \textbf{0.169}        & \textbf{0.341}          \\
\cline{2-8}
\multicolumn{1}{c|}{}                             & MoCo                &  \textbf{0.424}                  & 0.259                    & 0.174                   & 0.124                   & \textbf{0.169}        & 0.331         \\
\multicolumn{1}{c|}{}                             & SimCLR              & 0.423          & 0.259          & 0.175          & 0.126   & 0.166                 & 0.327    \\
\multicolumn{1}{c|}{}                             & SimCLR (w/ Lung Seg)&  0.415   & 0.255  & 0.174  & 0.127                    & 0.165                 & 0.331         \\
\hline
\multicolumn{1}{c|}{\multirow{6}{*}{MIMC-CXR}}    & Scratch  & 0.309                   & 0.193                   & 0.130                   & 0.093                   & 0.128                 & 0.274                   \\
\multicolumn{1}{c|}{}                             & AE       & 0.299                   & 0.184                   & 0.124                   & 0.088                   & 0.123                 & 0.270                   \\
\multicolumn{1}{c|}{}                             & Imagenet & 0.313                   & 0.196                   & 0.132                   & 0.095                   & 0.129                 & 0.275                   \\
\cline{2-8}
\multicolumn{1}{c|}{}                             & MoCo                &  0.309       & 0.190           & 0.127                   & 0.090                   & 0.126               & 0.269         \\
\multicolumn{1}{c|}{}                             & SimCLR              & 0.327          & 0.204       & 0.137          & 0.098   & 0.134                 & 0.278    \\
\multicolumn{1}{c|}{}                             & SimCLR (w/ Lung Seg)&  \textbf{0.335}   & \textbf{0.209}           & \textbf{0.140}                   & \textbf{0.100}                    & \textbf{0.136}                 & \textbf{0.279}         \\
\hline
\end{tabular}
\caption{Quantitative comparison among different encoders when using GRU as decoder. B-n, M, and R-L are short for BLEU-n, METEOR, and ROUGE-L, respectively.}
\label{Table:GRU_dif_encoder}
\end{table}

\subsection{Comparison with Previous Studies}
\begin{table}[!h]
\scriptsize
\centering
\setlength{\tabcolsep}{3pt}
\begin{tabular}{clccccc}
\hline
Dataset                    & \multicolumn{1}{c}{Method}                           & B-1           & B-2           & B-3           & B-4           & R-L           \\ \hline
\multirow{8}{*}{IU X-ray}  & Att2in$^\dagger$ \cite{rennie2017self}               & 0.399         & 0.249         & 0.172         & 0.126         & 0.321         \\
                           & AdaAtt$^\dagger$ \cite{lu2017knowing}                & 0.436         & 0.288         & 0.203         & 0.150         & 0.354         \\
                           & HRGR-Agent$^\dagger$ \cite{li2018hybrid}             & 0.438         & \textbf{0.298}& \textbf{0.208}& \textbf{0.151}& 0.322         \\
                           & CoAtt$^\dagger$ \cite{jing2017automatic}             & 0.455         & 0.288         & 0.205         & 0.154         & \textbf{0.369}\\ \cline{2-7} 
                           & Transformer \cite{vaswani2017attention}              & 0.420         & 0.256         & 0.175         & 0.127         & 0.326         \\
                           & Ours (SimCLR pretraining + Transformer)                          & \textbf{0.456}& 0.289         & 0.202         & 0.150         & 0.362         \\ \cline{2-7}
                           & R2Gen (Res50) \cite{chen2020generating}              & 0.455         & 0.293         & 0.210         & 0.160         & 0.360         \\ 
                           & Ours (MoCo pretraining + R2Gen)                                  & \textbf{0.466}& \textbf{0.303}& \textbf{0.218}& \textbf{0.165}& \textbf{0.361}\\ \hline
\multirow{5}{*}{MIMIC-CXR} & AdaAt$^\dagger$ \cite{lu2017knowing}                 & 0.299         & 0.185         & 0.124         & 0.088         & 0.266         \\
                           & Att2in$^\dagger$ \cite{rennie2017self}               & 0.325         & 0.203         & 0.136         & \textbf{0.096}& \textbf{0.276}\\ \cline{2-7}
                           & Transformer \cite{vaswani2017attention}              & 0.314         & 0.192         & 0.127         & 0.090         & 0.265         \\
                           & Ours (SimCLR pretraining + Transformer)                          & \textbf{0.343}& \textbf{0.208}& \textbf{0.137}& 0.095         & 0.260         \\ \cline{2-7}
                           & R2Gen (Res50) \cite{chen2020generating}              & 0.354         & 0.236         & 0.164         & 0.118         & \textbf{0.305}\\ 
                           & Ours (MoCo pretraining + R2Gen)                                  & \textbf{0.359}& \textbf{0.239}& \textbf{0.167}& \textbf{0.121}& 0.304         \\ \hline
\end{tabular}
\caption{Performance comparison with previous studies on the IU X-Ray dataset and MIMIC-CXR dataset. B-n and R-L are short for BLEU-n and ROUGE-L, respectively. $^\dagger$ indicates the results are quoted from the published paper. Note that we reproduce R2Gen \cite{chen2020generating} in both datasets with ResNet-50 architecture followed by their recommended setting.}
\label{Table:sota}
\end{table}

We compare our model with existing studies on the test set of both IU X-ray and MIMIC-CXR datasets in Table \ref{Table:sota}. Results show that the proposed method improves the performance with large margins without any extra information, such as pretraining on other datasets or keyword extraction. 

In addition, to show the feasibility of the CL pretrained encoder, we applied our MoCo pretrained encoder to the state-of-the-art model (R2Gen). Table \ref{Table:sota} shows the BLEU-1 score improved from 0.455 and 0.354 (R2Gen) to 0.466 and 0.359 (MoCo pretraining + R2Gen) when MoCo pretrained feature is used in IU X-ray and MIMIC-CXR dataset, respectively. In conclusion, results show that CL pretrained encoder gives representative image embeddings that can improve the performance of state-of-the-art methods.

\subsection{Quantitative Comparison between Encoders with and without Lung Segmentation}

In this section, we compare the effectiveness of lung segmentation augmentation. We use F1 scores on lung-related MeSH tags to evaluate whether lung segmentation gives more accurate descriptions regarding lung regions, as shown in Table \ref{Table:lung_keyword_f1}. The results show that lung segmentation would improve the accuracy of lung-related tags. This is because the encoder focuses more on lung regions and generates lung-specific features for the decoder. However, since the encoder cannot utilize features in the entire image, the overall performance of the generated descriptions would drop slightly, as shown in Table \ref{Table:Transformer_dif_encoder}.

\begin{table}[h]
\centering
\footnotesize
\setlength{\tabcolsep}{5pt}
\begin{tabular}{c|c|c}
\hline
Keyword       & CL w/ Lung Segmentation & CL w/o Lung Segmentation \\
\hline
pneumothorax  & \textbf{0.808}     & 0.752               \\
volume        & \textbf{0.191}     & 0.116               \\
effusion      & \textbf{0.789}     & 0.772               \\
calcification & \textbf{0.120}     & 0.030               \\
\hline
\end{tabular}
\caption{F1 score of lung-related keywords appeared in experiments w/ and w/o adding lung segmentation.}
\label{Table:lung_keyword_f1}
\end{table}




\subsection{Qualitative Comparison between Encoders with and without Lung Segmentation}
\begin{figure}[!h]
    \centering
    \includegraphics[width=1\linewidth]{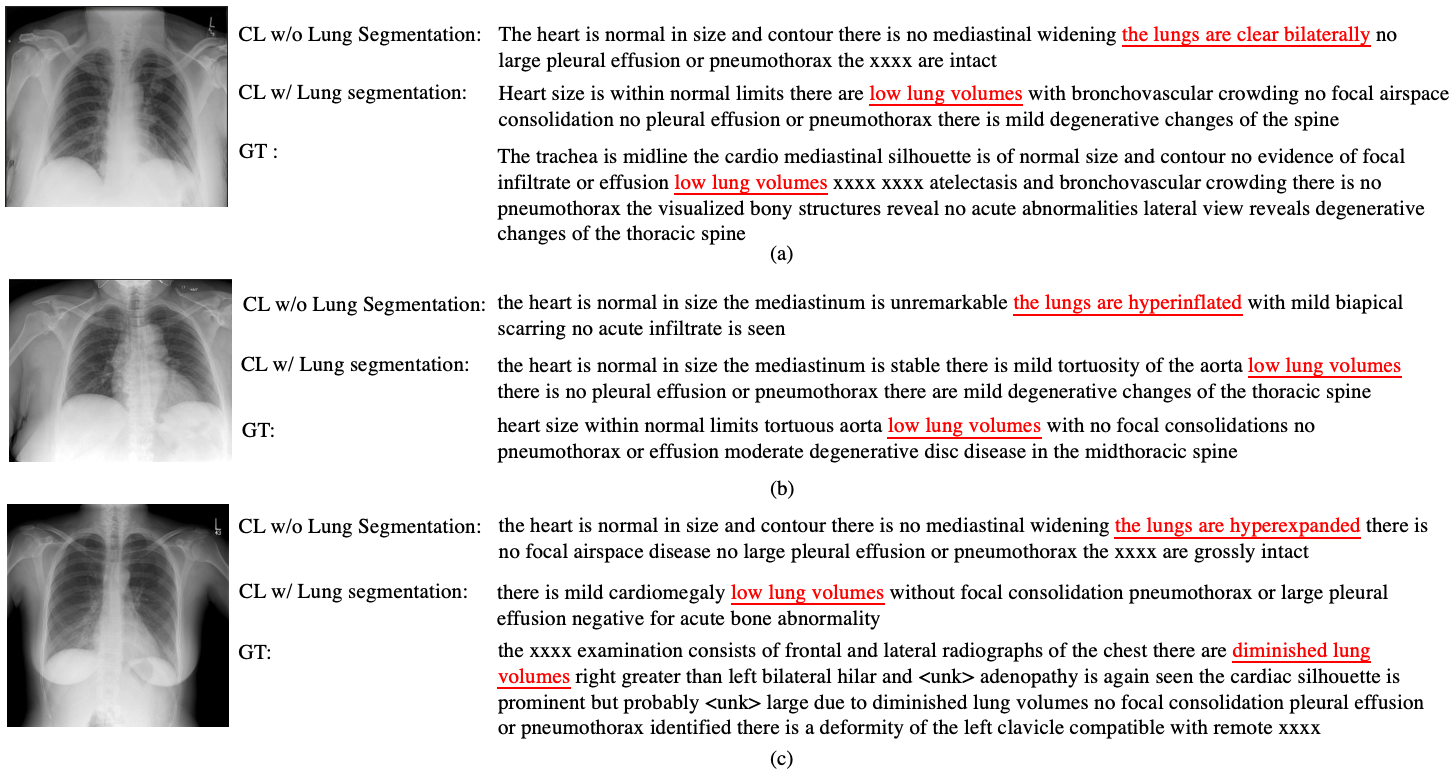}
    \caption{Qualitative comparison for whether lung segmentation is adopted for CL pretraining. With lung segmentation, the model can generate accurate descriptions about the lung, whereas the naive CL sometimes generates false descriptions about the lung volumes. GT stands for the ground truth description.}
    \label{fig:lm_comp}
\end{figure}

In this section, we discuss the qualitative comparison between encoders pretrained by using and not using lung segmentation-based augmentation.
Fig. \ref{fig:lm_comp} shows the ground truth and the predicted descriptions by different encoders. In case (a), the CL model only describes the clearness of the lungs, whereas with lung segmentation, the model can also point out low lung volumes which match the ground truth description. 
In case (b) and (c), the CL model falsely predicts an expanded (inflated) lungs. In contrast, the lung segmentation model and the ground truth describe a ``low lung volume''. This shows that lung segmentation aids the model in generating accurate descriptions, especially on the lung area.

\section{Conclusion}
In this work, we showed that the contrastive learning method provides high-quality image representation for chest X-ray report generation. Unlike most existing approaches, which pretrained the encoder with the dataset of different domains, our method successfully extracted the feature from the target dataset, which there is no domain bias. Furthermore, with lung segmentation included, the network generates more detailed and accurate descriptions regarding the lung regions. Experimental results on two datasets have shown that our framework improved the performance with a large margin from other pretraining strategies.

\bibliography{midl}

\section*{Acknowledgement}
We thank the National Center for High-performance Computing (NCHC) for providing computational and storage resources.




\end{document}